\ifdefined\XeTeXversion\else\ifdefined\directlua\else\pdfoutput=1\fi\fi
\documentclass[11pt]{article}
\usepackage[numbers]{thefinai}

\renewcommand{\thefinaiwordmark}[1][1.7em]{%
  \includegraphics[height=18pt]{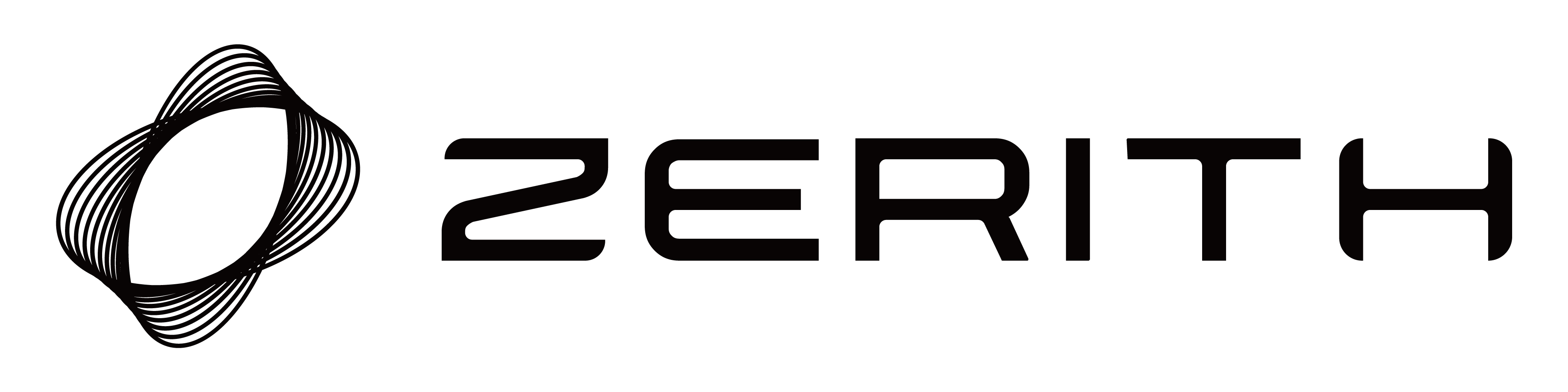}}
\usepackage{xurl}
\usepackage{needspace}
\usepackage{float}
\usepackage{caption}

\makeatletter
\def\bstctlcite{\@ifnextchar[{\@bstctlcite}{\@bstctlcite[@auxout]}}
\def\@bstctlcite[#1]#2{\@bsphack
  \@for\@citeb:=#2\do{%
    \edef\@citeb{\expandafter\@firstofone\@citeb}%
    \if@filesw\immediate\write\csname #1\endcsname{\string\citation{\@citeb}}\fi}%
  \@esphack}
\makeatother
\newcolumntype{C}{>{\centering\arraybackslash}X}

\newcommand{\vla}{\ensuremath{\pi_{\theta}}}
\newcommand{\pizero}{\ensuremath{\pi_{0.5}}}
\newcommand{\LLM}{\ensuremath{\mathcal{T}}}
\newcommand{\dataset}{\ensuremath{\mathcal{D}}}
\newcommand{\buffer}{\ensuremath{\mathcal{B}}}
\newcolumntype{Y}{>{\raggedright\arraybackslash}X}
\hypersetup{pdftitle={ESI: An Embodied Self-Improvement Flywheel for LLM-VLA Agents}}

\title{ESI: An Embodied Self-Improvement Flywheel\\for LLM--VLA Agents}
\date{November 11, 2026}
\reportlabel{Technical Report}
\author{%
  Yundi Cai\textsuperscript{1,2}\quad
  Zihan Wang\textsuperscript{2,3}\quad
  Chenxi Han\textsuperscript{2,3}\quad
  Yi Cheng\textsuperscript{2,3}\quad
  Houde Liu\textsuperscript{3}\quad
  Zilei Wang\textsuperscript{1}%
}
\affiliation[1]{University of Science and Technology of China}

\affiliation[2]{ZERITH Robotics}

\affiliation[3]{Tsinghua Shenzhen International Graduate School}
\metadata[Project]{\url{https://www.zerith.com/blog/esi/}}
\metadata[Code]{\ghlink{ZERITH-ZLab/ESI}}

\newcommand{\ESIHeaderBrand}{%
  \raisebox{-0.2\height}{%
    \includegraphics[
      height=12pt,
      trim=67.2bp 52.8bp 67.2bp 53.28bp,
      clip
    ]{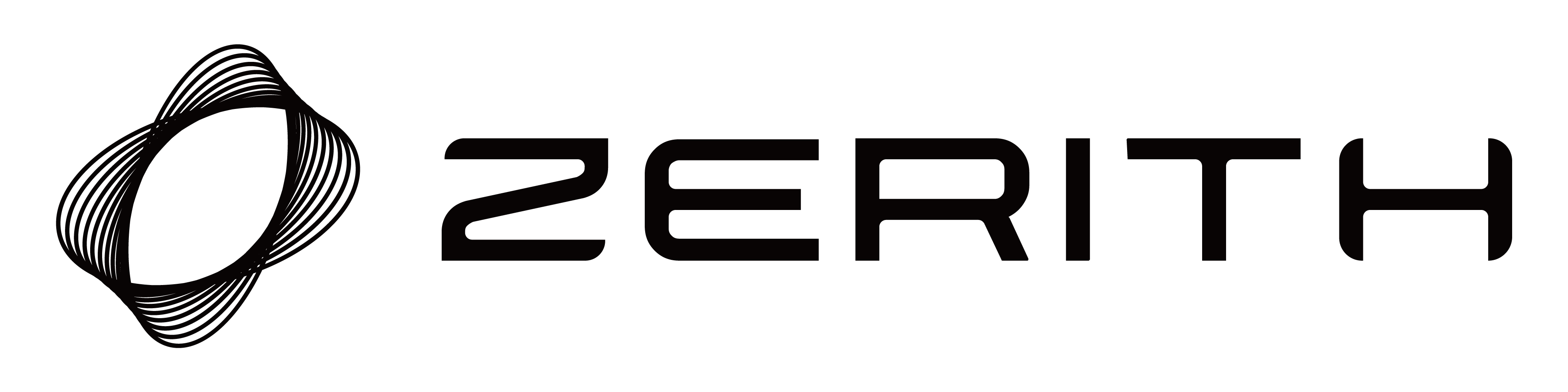}%
  }%
}

\newcommand{\ESIHeaderStyle}{%
  \fancyhf{}%
  \fancyhead[L]{\ESIHeaderBrand}%
  \fancyhead[R]{%
    \sffamily\fontsize{8}{10}\selectfont
    {\bfseries\color{finaiblue}ESI}%
    \hspace{0.65em}{\color{finaimuted}/}\hspace{0.65em}%
    {\color{finaisoft}TECHNICAL REPORT}%
  }%
  \fancyfoot[C]{\sffamily\footnotesize\color{finaisoft}\thepage}%
  \renewcommand{\headrulewidth}{0.2pt}%
  \renewcommand{\footrulewidth}{0pt}%
  \renewcommand{\headrule}{%
    {\color{finailine}%
     \hrule width\headwidth height\headrulewidth
     \vskip-\headrulewidth}%
  }%
}
\fancypagestyle{esi}{\ESIHeaderStyle}
\fancypagestyle{plain}{\ESIHeaderStyle}
\begin{document}
\bstctlcite{ESI:BSTcontrol}

\begin{abstract}
Independently pretrained language models and vision--language--action (VLA) policies provide complementary building blocks for embodied agents, but connecting them does not automatically produce coordinated behavior. The VLA may generate actions that diverge from the LLM's intended subtask and, upon regaining control after a correction, undo the promising state restored by the LLM. We introduce ESI (Embodied Self-Improvement), an automated flywheel that uses environmental interaction to align reasoning and action. ESI treats the modular agent as the unit of post-training, repeatedly refining the prompts that guide the LLM, the shared skill interface that expresses its intent, and the VLA parameters that realize that intent, then redeploying these components together. Our key insight is that collaboration itself can supply supervision for improving this interface. The LLM's subtask decisions provide semantic supervision for what the policy should execute, while its effective corrections and subsequent student continuations provide action-level supervision for recovery and handoff. The student also learns from its own effective execution, allowing locally useful experience to contribute even when the overall episode fails. 
On ten selected RoboDojo tasks, ESI achieves a mean Score of 62.03 and a mean success rate of 49.41\% using an LLM--VLA agent without generating task-specific control code. On both metrics, it outperforms the published reference results of all hybrid embodied agents, including the Agent+WAM system GO-3 Nano Preview (47.27 Score and 36.60\% success), as well as all compared code-based methods.
\end{abstract}

\maketitle

\begin{figure}[t]
  \centering
  \includegraphics[width=\linewidth]{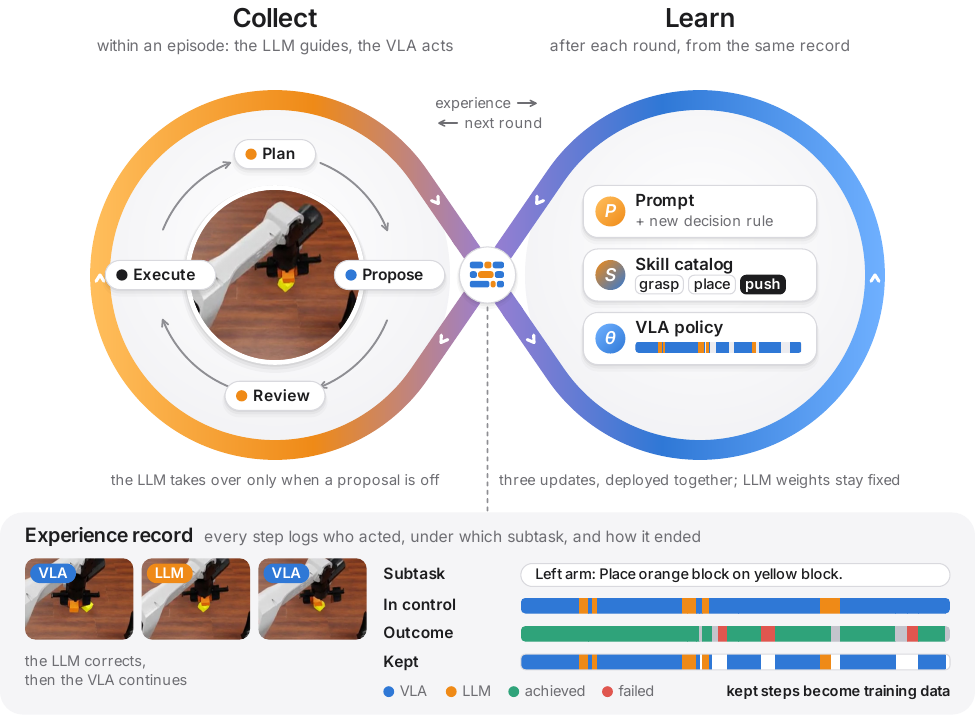}
  \caption{\textbf{The ESI collect--learn flywheel.} Within each episode, the LLM plans and reviews, the VLA proposes actions, and teacher corrections create states from which the VLA continues. The shared experience record links local intent, control source, and local execution outcomes. After each collection round, experience informs prompt and skill refinement, while retained effective actions provide VLA training data. The three updated components are deployed together in the next round; LLM weights remain fixed.}
  \label{fig:automation}
\end{figure}

\section{Introduction}
\label{sec:intro}

Robotics is entering a stage in which strong general-purpose reasoning models can participate directly in the control loop. They can interpret unfamiliar instructions, identify relevant objects, reason about intermediate goals, and revise a plan after observing a failure. At the same time, VLA policies provide a powerful substrate for generating continuous robot actions from images, language, and proprioception~\citep{black2025pi0,black2025pi05}. Agentic systems that combine these capabilities are a compelling route toward more general robots: a reasoning model decides what should happen next, and an action model provides the motor behavior needed to make it happen~\citep{ichter2023saycan,shi2025hirobot,gemini2025robotics15,galbot2026gptaspolicy}.

The central challenge is making this division of labor effective. Both components may be capable in isolation, yet their composition can fail at the interface. A task-finetuned VLA may recognize a familiar scene and reproduce a familiar motion while underusing the teacher's more specific instruction. Related work identifies visual shortcuts that allow scene information to dominate language constraints~\citep{lian2026langforce}. A teacher may correctly diagnose a bad approach, adjust the gripper, and restore a promising configuration, only for the student to undo that correction when control returns. These failures expose two closely related gaps: an \emph{intent gap} between the teacher's decision and the student's behavior, and a \emph{handoff gap} between the state produced by an intervention and the student's ability to continue.

Consider an illustrative packing task. The teacher selects the red object and asks the left arm to grasp it. The student instead approaches another object, or closes its gripper before reaching a suitable contact configuration. The teacher corrects the approach and gives control back. If the student has never learned to act from this corrected state under the intended subtask, it may repeat the original motion or abandon the target. The immediate problem can be solved with more teacher control, but the same coordination failure may recur in the next episode. A system that only intervenes repeatedly spends reasoning effort without necessarily improving its underlying action capability.

Our premise is that these interactions are valuable learning data. They identify where the deployed policy diverges from a selected goal, show how that divergence can be corrected, and expose the states from which execution must resume. We therefore treat the reasoning--action interface as a target of \emph{collaborative post-training}. The teacher remains an online planner, reviewer, and source of corrections. Their joint experience updates the teacher's guidance, the shared skill vocabulary, and the student policy so that future cooperation becomes more effective. This creates a path from episodic assistance to persistent capability.

This motivates a shift from model-level to \emph{system-level post-training}. In a modular embodied agent, decision guidance, the language interface, and motor behavior jointly determine what the system can accomplish. ESI makes their cooperation a learning target: interaction supplies evidence for updating each component, and their joint deployment determines the experience collected next. Through this process, the LLM's scene understanding and corrective judgments become supervision for persistent action capability.

Web-to-robot knowledge transfer~\citep{zitkovich2023rt2} and language-guided policy distillation~\citep{ha2023scalingup} illustrate how foundation models can support embodied learning. ESI provides a pathway for converting advances in general-purpose foundation models into executable embodied capabilities. This pathway has two stages. During deployment, the LLM contributes scene understanding, planning, and corrective decisions to the current agent. Through the flywheel, effective interaction becomes supervision that updates the action policy and guidance for subsequent rounds. A foundation model thus participates both in performing tasks and in producing the experience from which the embodied system learns.

We instantiate this idea in ESI (Embodied Self-Improvement), a configure-once flywheel built on the open-source GPT-as-Policy system~\citep{galbot2026gptaspolicy} (Figure~\ref{fig:automation}). During collection, the teacher selects a structured subtask, the VLA proposes an action chunk, and the teacher accepts useful motion or applies a correction. The resulting observations, instructions, reasoning, executed actions, and feedback support round-level updates of decision guidance, the shared skill catalog, and policy parameters. These components are deployed together in the next round, while the LLM's weights remain fixed.

The link from interaction to policy improvement is the selection of locally effective behavior. ESI retains successful subtasks from unsuccessful episodes and connects teacher corrections with valid student continuations on the executed timeline. Training under the corresponding local intent teaches the policy both what operation to perform and how to continue from states reached during collaboration. Episode-level review complements this action supervision by refining the teacher's guidance and consolidating recurring custom operations into reusable skills.

On the selected RoboDojo tasks, ESI scores higher than both individual component references on all ten tasks and exceeds both in success rate on nine. For packing objects into a box, ESI completes ten of twenty cases, against reported success rates of 0\% for GPT-6-Astra and 2.67\% for $\pi_{0.5}$. Section~\ref{sec:results} reports the full comparison.

This report develops three main contributions:
\begin{itemize}
  \item \textbf{System-level post-training for modular embodied agents.} We make the cooperation between independently pretrained reasoning and action models a target of adaptation, updating decision guidance, the shared skill interface, and motor execution through environmental interaction.
  \item \textbf{Interaction-grounded embodied distillation.} We connect semantic intent and corrective behavior to VLA learning through a shared control and supervision interface. The pipeline retains successful student behavior, effective corrections, and their continuations, including useful subtasks from unsuccessful episodes.
  \item \textbf{A framework for automated embodied self-improvement.} We automate experience-driven updates of decision prompts, a shared skill catalog, and VLA parameters, and return the updated components to the next deployment round.
\end{itemize}

\section{Related Work}
\label{sec:related}

\subsection{Reasoning models and language-conditioned robot policies}

SayCan grounds language-model planning in robot skills and their affordances~\citep{ichter2023saycan}, while Code as Policies uses generated programs to connect language instructions with perception and control APIs~\citep{liang2023codeaspolicies}. Hi Robot uses a hierarchical VLM--VLA architecture to interpret complex requests and situated feedback before executing the selected subtask~\citep{shi2025hirobot}. Gemini Robotics 1.5 develops an agentic architecture linking embodied reasoning with action generation~\citep{gemini2025robotics15}. More recently, GPT-as-Policy studies direct and hybrid uses of a general-purpose reasoning model across embodied tasks~\citep{galbot2026gptaspolicy}. These systems motivate a compositional view of robot intelligence. Our flywheel uses the experience produced by this composition to improve subsequent collaboration.

RT-2 transfers knowledge from web-scale vision--language pretraining into robot control through joint training on web and robot data~\citep{zitkovich2023rt2}. The $\pi_0$ family combines pretrained visual-language representations with flow-based action generation~\citep{black2025pi0}. \pizero{} extends this direction through heterogeneous training that includes semantic subtask information~\citep{black2025pi05}. LangForce studies visual shortcuts that weaken instruction following and explicitly models the contribution of language to action selection~\citep{lian2026langforce}. Such capabilities make a common language interface attractive, while task-specific post-training offers a practical way to connect a deployed teacher's instruction distribution to the student's motor behavior. Our focus is the interaction data generated by that particular teacher--student pairing.

\subsection{Learning from language guidance, deployment, and corrections}

Scaling Up and Distilling Down uses language-model-guided planning and robot motion planners to generate language-labeled trajectories, then distills them into a language-conditioned visuomotor policy~\citep{ha2023scalingup}. This connects language-guided data generation to persistent motor capability. ESI draws its training experience from online teacher--student collaboration, including effective interventions and student continuations.

Dataset aggregation makes the learner's visited states central to imitation learning~\citep{ross2011dagger}. Sirius learns from autonomous execution and human interventions using weighted behavioral cloning~\citep{liu2023sirius}. Yell At Your Robot uses human language corrections to improve a hierarchical robot's high-level policy~\citep{shi2024yayrobot}. RECAP combines demonstrations, autonomous experience, and expert interventions to improve VLA policies through advantage-conditioned learning~\citep{physicalintelligence2025recap}. Our system follows the broader principle that deployment should produce new learning opportunities. It uses a multimodal teacher to supply task decisions and corrections, and supervised flow-matching updates to absorb eligible mixed-control behavior. The shared subtask interface connects those updates to the teacher's intended operation.

\subsection{Experience-driven adaptation of modular agents}

Reflexion uses verbal feedback and episodic memory to improve subsequent decisions while keeping the language model's weights fixed~\citep{shinn2023reflexion}. Voyager accumulates reusable executable skills in Minecraft and refines programs using environmental feedback~\citep{wang2024voyager}. These works illustrate how interaction can improve an agent through its guidance and retained skills. In ESI, round-level review updates decision prompts and the shared instruction catalog alongside VLA post-training, connecting adaptation of the agent's context to adaptation of its motor policy.

DSPy represents language-model programs as composable modules and optimizes their pipelines against a task metric~\citep{khattab2024dspy}. GEPA uses reflection on execution traces to propose and evaluate prompt revisions~\citep{agrawal2026gepa}. These approaches make the configuration of a modular system an explicit optimization target. ESI brings that perspective into an embodied learning loop, where prompts, the shared skill interface, and motor-policy parameters adapt through environmental interaction.

FastHarness uses accumulated experience to schedule reviews of a frozen VLA, retaining recovery knowledge while reducing unnecessary supervision~\citep{lin2026fastharness}. CaPtain studies post-training of code-as-policy agents through supervised learning and multi-turn reinforcement learning~\citep{fan2026captain}. Together, these directions illustrate several places where an embodied agent can improve: its supervision strategy, its reasoning policy, and its motor policy. Our framework centers on learning the cooperation between an online reasoning teacher and a trainable VLA. Review scheduling and reasoning-model adaptation can be integrated as complementary branches.

\section{ESI: System-Level Post-Training for Embodied Agents}
\label{sec:perspective}

\subsection{Three coupled adaptation targets}

ESI treats the deployed agent as a composition $\mathcal{A}_r=(\LLM;P_r,S_r,\theta_r)$ at collection round $r$. Here $\LLM$ is a fixed pretrained language model, $P_r$ contains its system and task guidance, $S_r$ defines the shared skill catalog and argument schema, and $\theta_r$ contains the VLA parameters. Environmental interaction yields an experience buffer $\buffer_r$. The flywheel uses this shared evidence to update the prompt, skill catalog, and policy for the next collection round.

These objects address different sources of coordination failure. Prompt evolution changes how the teacher plans, reviews action, and responds to difficulty. Skill evolution turns recurring operations into reusable descriptions that make the teacher's intent explicit. Policy learning changes how the robot realizes that intent from the states encountered during collaboration. The skill catalog defines the shared language used by a single VLA policy; each entry describes an operation rather than a separately trained controller.

The coupling is carried by subsequent interaction: changes in guidance and skill descriptions alter the operations collected, while policy updates change which operations still require teacher intervention.

\subsection{Learning the interface between reasoning and action}

Let $g$ denote the episode-level instruction, $o_t$ the available observation including images and robot state, and $h_t$ the interaction history. A multimodal teacher $\LLM$ uses this context to select a local intent and a structured subtask $z_t$. The student policy $\vla$ maps the observation and a rendered language condition $c_t$ to an action chunk. The teacher then decides which prefix to execute or how to correct it.

The two models communicate through language, action proposals, and environmental feedback. Having compatible input and output formats makes this interaction possible, but reliable cooperation also requires behavioral alignment. We focus on two properties:

\paragraph{Intent following} The student should translate the teacher's selected object, acting arm, destination, relation, and orientation into appropriate motion. A meaningful subtask instruction should change behavior when the current goal changes.
\paragraph{Continuation after intervention} The student should preserve useful changes made by the teacher and continue toward the active goal. A correction that improves contact geometry or object arrangement should become a viable starting point for subsequent student execution.

These properties are coupled. The same visual state may support several plausible actions, so a handoff requires both a recoverable physical configuration and an unambiguous goal. Conversely, a correct instruction has limited value when the student cannot act from the state produced by the preceding correction. The learning unit is therefore the combination of \emph{observation, intended operation, and effective executed behavior}.

\subsection{System-level post-training and embodied knowledge transfer}
\label{sec:knowledge-transfer}

We interpret the transfer from teacher-guided interaction to policy learning as \emph{interaction-grounded embodied distillation}. It connects the teacher's embodied knowledge to persistent policy capability at two complementary levels.

\paragraph{Semantic-level transfer}
The LLM applies its general knowledge to interpret a scene, identify relevant objects and relations, and select the next operation. A structured subtask turns this reasoning into an explicit local goal. For example, within a packing task, ``Left arm: Grasp red block'' identifies what the current action segment should accomplish. Training on effective behavior under this condition teaches the VLA to realize the selected intent. Local supervision connects each action segment to its immediate purpose, while the teacher continues to organize those operations into the broader task.

\paragraph{Action-level transfer}
The LLM also expresses embodied understanding through its judgment of proposed motion and its corrections during execution. It can repair an approach, adjust an object orientation, or restore a useful grasp. Recording the effective actions and subsequent continuation turns these judgments into behavioral supervision. The policy learns from the physical states reached during collaboration, including states created by teacher intervention. This gives a successful correction value beyond the episode in which it occurs.

\paragraph{One interface for control and learning}
The language used to request behavior online is also the language used to learn that behavior offline. The shared skill interface therefore connects both levels of transfer: it conveys the teacher's selected intent and conditions the effective actions retained for training. As custom operations are consolidated into reusable skills, the interface acquires new ways to express recurring demands. ESI uses this connection to carry experience from online cooperation into subsequent policy execution.

\subsection{From online feedback to round-level updates}

\begin{figure}[t]
\centering
\includegraphics[width=\linewidth]{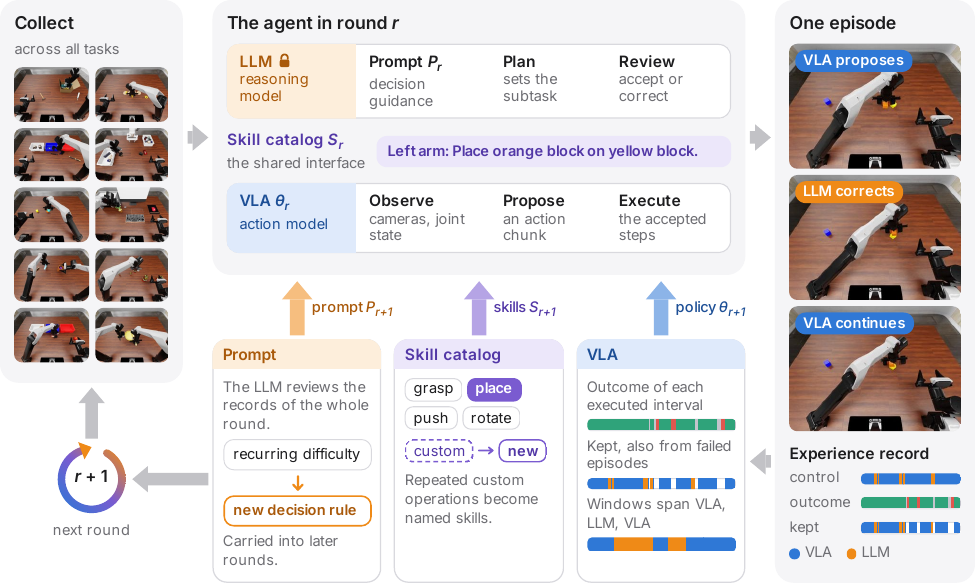}
\caption{\textbf{ESI architecture and round-level updates.} The agent combines a fixed LLM with decision guidance $P_r$, a shared skill catalog $S_r$, and a VLA policy $\theta_r$. Interaction across tasks produces records for three update paths: reviewing difficulties to refine guidance, consolidating recurring custom operations into reusable skills, and constructing policy supervision from locally effective execution. Prompt and skill review uses the interaction records, while VLA training uses the retained effective actions. Training windows preserve continuity across teacher--student handoffs and retain useful behavior from unsuccessful episodes. Updated components form the next-round agent.}
\label{fig:overview}
\end{figure}

ESI operates at two complementary levels. Within an episode, the teacher and VLA use fresh observations to plan, review, correct, and continue execution. Across collection rounds, the system consolidates experience into updated prompts, skills, and policy parameters. Episode summaries also provide task memory for subsequent attempts. Figure~\ref{fig:overview} expands the current-round agent, its interaction record, and the three update paths that form the next-round agent.

\section{Online Collaboration and Experience Collection}
\label{sec:online}

\subsection{A shared subtask language}
\label{sec:schema}

The teacher expresses the current local goal through a shared subtask interface. Each subtask identifies the operation, acting arm, manipulated object, and relevant target or orientation. The system renders this representation as a concise instruction appended to the original task and supplies it to the VLA. We refer to this task-plus-local-instruction input as \emph{subtask conditioning}.

\begin{promptbox}[Illustrative student input]
Put the red block in the tray.\\
Current subtask: Left arm: Grasp red block.
\end{promptbox}

The same expression of local intent is used both to request actions online and to train the policy on the corresponding execution experience. The catalog provides reusable operations such as grasping, placing, and rotating, while a \texttt{custom} option expresses behavior beyond its current vocabulary. Recurring custom operations provide evidence for subsequent skill expansion. Appendix~\ref{app:schema} gives the interface fields and representative instructions.

\subsection{Plan, propose, review, and execute}

Each decision cycle begins with the current observation and task context. The teacher selects a subtask, and the VLA proposes an action chunk conditioned on the observation and the resulting instruction. The teacher reviews which actions to execute using images, robot state, recent feedback, and the proposal's forward-kinematics trajectory.

The teacher can accept a useful segment, correct its motion or gripper behavior, or directly specify a recovery motion. Execution produces fresh observations and feedback for the next planning and review cycle. The VLA supplies continuous actions, while the teacher adapts the control decision to the intended operation and observed task progress.

Teacher corrections and student continuations together form the executed trajectory. ESI records these handoffs so that subsequent learning can use both the states reached through teacher recovery and the student's behavior after control returns.

\subsection{A shared record of experience}
\label{sec:buffer}

The shared experience record links observations, local instructions, proposed actions, teacher reasoning, control sources, executed actions, feedback, and outcomes. Episode summaries organize the difficulties encountered and the strategies attempted, and can serve as task memory for subsequent attempts.

This record supports three complementary updates. Summaries and detailed interaction evidence inform decision guidance; recurring custom operations inform skill consolidation; and executed trajectories with local outcome annotations provide policy supervision. The next section describes how ESI uses these sources to update the agent jointly across rounds.

\section{Learning from Interaction: Prompt, Skill, and Policy Updates}
\label{sec:learning}

After each collection round, ESI updates decision guidance, the shared skill catalog, and the action policy from the same experience record. Prompt updates shape planning and review, skill updates shape the expression of local intent, and policy updates learn to realize that intent from effective execution. The updated components enter the next interaction round together and generate new evidence for learning.

\subsection{Experience-driven prompt evolution}
\label{sec:prompt-evolution}

Reflective prompt optimization uses execution traces to turn observed difficulties into revised instructions~\citep{agrawal2026gepa}. ESI reviews episode summaries, outcomes, and teacher-control histories to identify recurring manipulation and recovery difficulties. It then examines relevant reasoning records, previous guidance, and selected images to connect those difficulties to the surrounding decisions and execution conditions.

The resulting revisions update reusable manipulation guidance, recovery rules, and task-specific strategies. An independent reviewer checks whether the proposals are supported by the recorded evidence and preserve productive existing behavior. Candidates requiring further evidence return for refinement; accepted guidance is automatically published for the next collection round. This process updates the teacher's decision context while keeping the LLM's weights fixed.

Prompt evolution turns difficulties encountered in individual interactions into reusable guidance for later attempts. Appendix~\ref{app:prompt-review} details the review criteria and a recorded proposal, including the evidence used to determine whether it should be published.

\subsection{From custom operations to reusable skills}
\label{sec:skill-evolution}

The shared skill catalog evolves from operations that arise during collection. When the teacher frequently uses custom descriptions, the system reviews the associated instructions, reasoning, images, and execution context to identify recurring behavior that can be expressed through a stable, reusable skill description.

Suitable operations are consolidated into the catalog for subsequent rounds. This gives the teacher a consistent way to request an operation that previously required ad hoc wording and gives the VLA a reusable language condition for that intent. The \texttt{custom} option remains available for behavior beyond the revised catalog, exposing further opportunities for expansion.

Prompt evolution changes how the teacher decides and reviews; skill evolution changes how the selected operation is conveyed to the policy. Together, they shape the instruction distribution and interaction experience available for policy learning in subsequent rounds.

\subsection{Constructing policy supervision from interaction}
\label{sec:training}

\paragraph{Aligning local intent with executed behavior}
The pipeline aligns executed actions with their preceding observations and corresponding local intent. An LLM annotator uses the textual interaction record and post-action feedback to assess local outcomes and instruction--action consistency. Annotation preserves instructions supplied online and assigns local instructions when collection precedes online subtask conditioning. The resulting examples pair observations and intended operations with the motion actually executed. Reasoning and outcome evidence guide annotation and selection; the VLA receives images, robot state, and the task instruction with its applicable subtask condition.

\paragraph{Selecting locally effective behavior}
Selection follows local effectiveness and consistency with the intended operation. ESI retains productive student actions, effective teacher corrections, and preparatory actions whose progress is supported by subsequent feedback. Failed, uncertain, or inconsistent segments are excluded from direct action supervision. Selection is independent of the episode's final outcome: a successful grasp, transport, or correction can remain useful even if a later operation prevents task completion.

\paragraph{Connecting corrections and continuations}
Retained execution is organized into temporally continuous action windows. A switch between student and teacher control does not itself break a window, allowing a correction and the student's subsequent continuation to contribute jointly to supervision. These examples teach both how useful corrected states are reached and how execution continues from them. We refer to this action-level mechanism as \emph{intervention-to-policy distillation}. Supervision follows the contiguous valid portion of each window and stops when subsequent actions are no longer eligible.

Each window uses its starting observation and language condition. Near an operation transition, the pipeline can fall back to the task instruction alone, relaxing the outgoing local constraint. Appendix~\ref{app:annotation}--\ref{app:training-example} specifies the annotation, window construction, masking, and condition-transition rules. A recorded example there illustrates how a teacher correction and student continuation remain usable even when the episode ultimately fails.

\subsection{Policy updates with experience replay}
\label{sec:replay}

Each training round incorporates new interaction while retaining experience from earlier rounds. Let $\mu_r$ denote the sampling distribution over eligible training windows collected in round $r$, and let $q_r$ denote the distribution used for policy training after that round. The replay rule is recursive:
\begin{equation}
q_1=\mu_1,\qquad
q_r=\alpha\mu_r+(1-\alpha)q_{r-1},\quad r\geq2.
\label{eq:replay}
\end{equation}
We use $\alpha=0.6$: the current round receives 60\% of the sampling mass and the recursively organized history receives 40\%. Earlier rounds retain their relative proportions within that history. The rule applies to any number of collection rounds, allowing new experience to enter the training distribution without discarding previously acquired behavior. Within each round, sampling balances tasks and cases and gives greater weight to teacher-controlled window starts. Appendix~\ref{app:replay} details the resulting round weights and the within-round sampling procedure.

We instantiate the student with a \pizero{} policy~\citep{black2025pi05} using the OpenPI/JAX implementation\footnote{OpenPI: \url{https://github.com/Physical-Intelligence/openpi}.} and retain its native flow-matching objective~\citep{lipman2023flowmatching,black2025pi0}. Each round initializes from the preceding policy and updates the full VLA on the replay distribution. Action loss is applied only to the retained valid execution within each window; invalid and padded positions are excluded. Appendix~\ref{app:objective}--\ref{app:training-config} gives the full objective and training settings.

\subsection{Joint redeployment across rounds}
\label{sec:flywheel}

Let $\mathcal{A}_r=(\LLM;P_r,S_r,\theta_r)$ denote the agent deployed in round $r$, with fixed LLM weights $\LLM$, decision guidance $P_r$, a shared skill catalog $S_r$, and VLA parameters $\theta_r$. Its interaction record $\buffer_r$ supplies evidence for prompt and skill updates and yields the policy-training data $\dataset_r$:
\begin{equation}
\begin{aligned}
\buffer_r &\leftarrow \operatorname{Collect}(\mathcal{A}_r),\\
P_{r+1} &\leftarrow \operatorname{ReviewAndRefine}(\LLM,P_r,\buffer_r),\\
S_{r+1} &\leftarrow \operatorname{ConsolidateSkills}(\LLM,S_r,\buffer_r),\\
\dataset_r &\leftarrow \operatorname{AnnotateAndSelect}(\buffer_r),\\
\theta_{r+1} &\leftarrow \operatorname{PostTrain}(\theta_r,\dataset_{1:r};q_r),\\
\mathcal{A}_{r+1} &\leftarrow (\LLM;P_{r+1},S_{r+1},\theta_{r+1}).
\end{aligned}
\label{eq:flywheel}
\end{equation}
Here $\dataset_{1:r}$ contains the eligible execution windows accumulated through round $r$. The updated guidance and skills change how the teacher decides and expresses intent, while the updated VLA changes the actions it can execute and the corrections it requires. Their joint deployment therefore changes the experience collected next and supplies evidence for further adaptation.

This collect--update--redeploy cycle repeats across rounds, with task memory providing episode-level context. It connects online interaction, experience consolidation, and joint deployment into a continuing process of system-level improvement.

\section{Experiments}
\label{sec:results}

\subsection{Experimental setup}

\paragraph{System instantiation}
\label{sec:implementation}
The current system pairs GPT-6 Astra with a RoboDojo-adapted \pizero{} policy in simulated dual-arm ARX X5 manipulation. The student consumes a high camera view, two wrist-camera views, robot state, and language. Its action representation contains six joint targets and one gripper command per arm. The teacher operates through the GPT-as-Policy execution tools and reviews student proposals within the same episode context~\citep{galbot2026gptaspolicy}.

RoboDojo provides a diverse manipulation environment~\citep{chen2026robodojo}. The project's collection suite selects ten tasks involving packing, semantic sorting, number arrangement, sequential manipulation, and language-conditioned object organization. Its three-round configuration specifies 200 episodes per round, with cumulative pools of 200, 400, and 600 episodes. The three-round policy-learning configuration underlying the reported evaluation uses a fixed prompt and skill catalog across its VLA updates. The first round bootstraps subtask labels offline; subsequent rounds use online subtask conditioning and preserve those conditions during annotation. Evaluation uses the task and case selection described below.

\paragraph{Task and case selection} We adapt the ten-task suite of GPT-as-Policy~\citep{galbot2026gptaspolicy}, replacing several easier tasks with more demanding manipulation tasks. The main evaluation schedules ten tasks with twenty cases each, totaling 200 cases. Cases follow benchmark order: tasks with standard and random splits contribute equally from each split (ten each here), while single-list tasks use the first twenty cases. The component study follows the same balanced sampling rule with ten cases per task, totaling 100 cases. Aggregation details are consolidated in Appendix~\ref{app:additional-eval}.

\paragraph{Metrics and comparison} Score measures task progress on a 0--100 scale, and success rate (SR) measures full task completion, following RoboDojo~\citep{chen2026robodojo}. For each system, we report the arithmetic mean of their ten task-level Scores and SRs. The reference values come from the RoboDojo leaderboard\footnote{RoboDojo benchmark and leaderboard: \url{https://robodojo-benchmark.com/}.} for the same task names, using each entry's reported case set. The first three model columns in Table~\ref{tab:main-results} compare the deployed ESI system with its standalone components.

\paragraph{Matched system baseline} We additionally evaluate \emph{Baseline (task-only)}, a GPT-as-Policy agent with the original RoboDojo-adapted VLA, task-only policy conditioning, and baseline decision guidance. ESI uses the round-three checkpoint, structured subtask conditioning, and task guidance. Both use GPT-6-Astra with xhigh reasoning, the same environment code and matched layouts and scene seeds, the same motion executor, and identical native task step budgets. Each evaluation uses a fixed agent configuration and disables cross-episode task memory. The comparison measures the deployed systems as a whole; execution configuration is documented in Appendix~\ref{app:additional-eval}. These layouts are drawn from the flywheel's collection suite.

Both agents use the same ten-task, 200-case evaluation schedule. Performance uses task means to match the leaderboard table; control statistics use episode means of recorded teacher actions. A teacher intervention decision is an action-review decision that results in at least one executed teacher correction (edit or EEF action); it is not a count of LLM calls. Teacher-executed steps count the resulting environment actions. Tables abbreviate the decision metric as ``Interventions. The shared-success comparison uses the cases completed by both agents.

\subsection{Overall task performance}

\begin{table}[!htbp]\centering
\caption{\textbf{Manipulation on ten selected RoboDojo tasks.} Cells report Score / SR (\%). Blue shading increases with Score from 10 (white) to 95 (darkest blue). The final row weights tasks equally. GO-3 Nano Preview is listed as Agent+WAM on the leaderboard.}
\label{tab:main-results}
\begingroup\color{black}\setlength{\tabcolsep}{1pt}\renewcommand{\arraystretch}{1.48}\fontsize{7.0}{8.5}\selectfont
\makebox[\linewidth][c]{\begin{tabularx}{\dimexpr\paperwidth-32pt\relax}{l*{11}{>{\centering\arraybackslash}X}}\toprule
 & \multicolumn{3}{c}{ESI and its components} & \multicolumn{8}{c}{RoboDojo reference methods}\\\cmidrule(lr){2-4}\cmidrule(lr){5-12}
Task & \mbox{ESI} & \mbox{$\pi_{0.5}$} & \shortstack{GPT-6\\Astra} & \shortstack{GO-3 Nano\\Preview} & \mbox{HexaShell} & \mbox{WorldScape} & \mbox{EmbodiedEvo} & \mbox{WALL-OSS} & \mbox{VPP2} & \mbox{PhysicalRSI} & \mbox{Awomo-0.5}\\\midrule
Pack objects & \cellcolor[HTML]{74B5E7}\mbox{65.75/50.00\%} & \cellcolor[HTML]{EDF6FC}\mbox{19.10/2.67\%} & \cellcolor[HTML]{FCFDFF}\mbox{11.60/0.00\%} & \cellcolor[HTML]{FFFFFF}\mbox{6.20/0.00\%} & \cellcolor[HTML]{D1E6F6}\mbox{33.10/16.00\%} & \cellcolor[HTML]{D3E7F6}\mbox{31.50/4.00\%} & \cellcolor[HTML]{F7FBFE}\mbox{14.10/0.00\%} & \cellcolor[HTML]{D0E5F6}\mbox{33.23/8.66\%} & \cellcolor[HTML]{D4E8F7}\mbox{30.43/6.67\%} & \cellcolor[HTML]{EAF4FC}\mbox{20.67/2.67\%} & \cellcolor[HTML]{E3F0FA}\mbox{23.80/6.00\%}\\
Sweep blocks & \cellcolor[HTML]{FFFFFF}\mbox{10.00/10.00\%} & \cellcolor[HTML]{FFFFFF}\mbox{0.00/0.00\%} & \cellcolor[HTML]{FFFFFF}\mbox{4.00/4.00\%} & \cellcolor[HTML]{FFFFFF}\mbox{0.00/0.00\%} & \cellcolor[HTML]{FFFFFF}\mbox{8.00/8.00\%} & \cellcolor[HTML]{FFFFFF}\mbox{0.00/0.00\%} & \cellcolor[HTML]{FFFFFF}\mbox{2.00/2.00\%} & \cellcolor[HTML]{FFFFFF}\mbox{9.33/9.33\%} & \cellcolor[HTML]{FEFEFF}\mbox{10.66/10.66\%} & \cellcolor[HTML]{FFFFFF}\mbox{0.00/0.00\%} & \cellcolor[HTML]{FFFFFF}\mbox{0.67/0.67\%}\\
Largest number & \cellcolor[HTML]{4B96D9}\mbox{83.25/75.00\%} & \cellcolor[HTML]{FFFFFF}\mbox{2.77/0.67\%} & \cellcolor[HTML]{6DB0E5}\mbox{68.70/60.00\%} & \cellcolor[HTML]{6FB1E5}\mbox{67.80/62.00\%} & \cellcolor[HTML]{D5E8F7}\mbox{30.20/22.00\%} & \cellcolor[HTML]{69ADE3}\mbox{71.00/62.00\%} & \cellcolor[HTML]{B2D5F0}\mbox{48.30/32.00\%} & \cellcolor[HTML]{C9E1F4}\mbox{38.23/24.00\%} & \cellcolor[HTML]{FFFFFF}\mbox{8.30/2.00\%} & \cellcolor[HTML]{EFF7FD}\mbox{18.17/11.33\%} & \cellcolor[HTML]{EFF7FD}\mbox{17.90/8.67\%}\\
Sort nesting dolls & \cellcolor[HTML]{EBF5FC}\mbox{20.00/20.00\%} & \cellcolor[HTML]{FFFFFF}\mbox{5.33/5.33\%} & \cellcolor[HTML]{EFF7FD}\mbox{18.00/18.00\%} & \cellcolor[HTML]{C1DDF2}\mbox{44.00/44.00\%} & \cellcolor[HTML]{A2CEEE}\mbox{52.00/52.00\%} & \cellcolor[HTML]{FFFFFF}\mbox{8.00/8.00\%} & \cellcolor[HTML]{F7FBFE}\mbox{14.00/14.00\%} & \cellcolor[HTML]{B3D6F0}\mbox{48.00/48.00\%} & \cellcolor[HTML]{EEF6FC}\mbox{18.67/18.67\%} & \cellcolor[HTML]{FFFFFF}\mbox{8.00/8.00\%} & \cellcolor[HTML]{FCFEFF}\mbox{11.34/11.34\%}\\
Classify objects & \cellcolor[HTML]{3D89D0}\mbox{89.75/85.00\%} & \cellcolor[HTML]{E1EFFA}\mbox{24.67/12.67\%} & \cellcolor[HTML]{6DB0E5}\mbox{69.00/56.00\%} & \cellcolor[HTML]{549DDD}\mbox{79.70/72.00\%} & \cellcolor[HTML]{4591D7}\mbox{85.30/78.00\%} & \cellcolor[HTML]{3985CD}\mbox{91.90/90.00\%} & \cellcolor[HTML]{CDE4F5}\mbox{35.50/16.00\%} & \cellcolor[HTML]{79B8E8}\mbox{63.43/54.00\%} & \cellcolor[HTML]{7DBBE9}\mbox{61.63/53.33\%} & \cellcolor[HTML]{92C6EC}\mbox{55.83/40.00\%} & \cellcolor[HTML]{D3E7F7}\mbox{31.20/20.67\%}\\
Fill pen holder & \cellcolor[HTML]{D0E5F6}\mbox{33.25/5.00\%} & \cellcolor[HTML]{E4F1FA}\mbox{23.23/7.33\%} & \cellcolor[HTML]{FFFFFF}\mbox{5.00/0.00\%} & \cellcolor[HTML]{EFF7FD}\mbox{17.80/4.00\%} & \cellcolor[HTML]{FFFFFF}\mbox{6.10/0.00\%} & \cellcolor[HTML]{D2E6F6}\mbox{32.30/8.00\%} & \cellcolor[HTML]{FFFFFF}\mbox{0.90/0.00\%} & \cellcolor[HTML]{BFDCF2}\mbox{45.23/25.33\%} & \cellcolor[HTML]{AAD2EF}\mbox{50.07/29.33\%} & \cellcolor[HTML]{E7F3FB}\mbox{21.87/3.33\%} & \cellcolor[HTML]{B2D5F0}\mbox{48.27/26.00\%}\\
Organize table & \cellcolor[HTML]{78B8E8}\mbox{63.75/10.00\%} & \cellcolor[HTML]{E4F1FA}\mbox{23.33/0.00\%} & \cellcolor[HTML]{C8E0F4}\mbox{39.50/0.00\%} & \cellcolor[HTML]{62A8E1}\mbox{74.00/16.00\%} & \cellcolor[HTML]{F7FBFE}\mbox{14.00/0.00\%} & \cellcolor[HTML]{7DBBE9}\mbox{61.50/22.00\%} & \cellcolor[HTML]{CEE4F5}\mbox{35.00/2.00\%} & \cellcolor[HTML]{6AAEE4}\mbox{70.17/17.33\%} & \cellcolor[HTML]{79B8E8}\mbox{63.17/7.33\%} & \cellcolor[HTML]{DBEBF8}\mbox{27.33/4.00\%} & \cellcolor[HTML]{ADD3F0}\mbox{49.50/9.33\%}\\
Imitate sorting & \cellcolor[HTML]{337FC8}\mbox{97.06/94.12\%} & \cellcolor[HTML]{FFFFFF}\mbox{1.60/0.00\%} & \cellcolor[HTML]{85BFEB}\mbox{58.90/36.00\%} & \cellcolor[HTML]{DCECF8}\mbox{27.00/22.00\%} & \cellcolor[HTML]{79B8E8}\mbox{63.20/42.00\%} & \cellcolor[HTML]{C1DDF2}\mbox{44.10/32.00\%} & \cellcolor[HTML]{5EA5E0}\mbox{75.80/60.00\%} & \cellcolor[HTML]{88C1EB}\mbox{58.17/50.67\%} & \cellcolor[HTML]{FFFFFF}\mbox{2.73/0.00\%} & \cellcolor[HTML]{FFFFFF}\mbox{1.10/0.00\%} & \cellcolor[HTML]{F4F9FD}\mbox{15.60/9.33\%}\\
Stack by language & \cellcolor[HTML]{5DA5E0}\mbox{76.00/70.00\%} & \cellcolor[HTML]{FFFFFF}\mbox{1.73/0.00\%} & \cellcolor[HTML]{B3D6F0}\mbox{48.00/40.00\%} & \cellcolor[HTML]{4F99DA}\mbox{81.60/80.00\%} & \cellcolor[HTML]{337FC8}\mbox{96.40/96.00\%} & \cellcolor[HTML]{58A1DE}\mbox{78.00/74.00\%} & \cellcolor[HTML]{A4CEEE}\mbox{51.60/50.00\%} & \cellcolor[HTML]{4591D7}\mbox{85.33/83.33\%} & \cellcolor[HTML]{F8FBFE}\mbox{13.60/12.67\%} & \cellcolor[HTML]{FFFFFF}\mbox{0.00/0.00\%} & \cellcolor[HTML]{F6FAFE}\mbox{14.67/11.33\%}\\
Classify by language & \cellcolor[HTML]{4F99DA}\mbox{81.50/75.00\%} & \cellcolor[HTML]{FFFFFF}\mbox{0.60/0.00\%} & \cellcolor[HTML]{BCDAF1}\mbox{46.00/30.00\%} & \cellcolor[HTML]{61A7E1}\mbox{74.60/66.00\%} & \cellcolor[HTML]{559EDD}\mbox{79.40/72.00\%} & \cellcolor[HTML]{71B3E6}\mbox{66.80/58.00\%} & \cellcolor[HTML]{89C1EB}\mbox{58.00/40.00\%} & \cellcolor[HTML]{FFFFFF}\mbox{7.27/0.67\%} & \cellcolor[HTML]{FFFFFF}\mbox{5.27/2.00\%} & \cellcolor[HTML]{E9F4FC}\mbox{20.93/6.00\%} & \cellcolor[HTML]{FFFFFF}\mbox{8.73/3.33\%}\\
\midrule Task mean & \cellcolor[HTML]{7CBAE9}\mbox{62.03/49.41\%} & \cellcolor[HTML]{FFFFFF}\mbox{10.24/2.87\%} & \cellcolor[HTML]{CBE3F5}\mbox{36.87/24.40\%} & \cellcolor[HTML]{B6D7F1}\mbox{47.27/36.60\%} & \cellcolor[HTML]{B8D8F1}\mbox{46.77/38.60\%} & \cellcolor[HTML]{B1D5F0}\mbox{48.51/35.80\%} & \cellcolor[HTML]{D0E5F6}\mbox{33.52/21.60\%} & \cellcolor[HTML]{BCDAF2}\mbox{45.84/32.13\%} & \cellcolor[HTML]{DDEDF9}\mbox{26.45/14.27\%} & \cellcolor[HTML]{F0F8FD}\mbox{17.39/7.53\%} & \cellcolor[HTML]{E6F2FB}\mbox{22.17/10.67\%}\\
\bottomrule\end{tabularx}}\endgroup
\end{table}

\paragraph{Overall performance} Averaged equally across the ten selected tasks, ESI reaches 62.03 Score and 49.41\% SR, exceeding all reference methods in Table~\ref{tab:main-results} on both metrics. The strongest reference Score is 48.51 from WorldScape-R$^2$SI, while the strongest reference SR is 38.60\% from HexaShell-Preview; ESI exceeds these by 13.52 Score and 10.81 percentage points, respectively. GO-3 Nano Preview, listed as Agent+WAM on the RoboDojo leaderboard, reaches 47.27 Score and 36.60\% SR on the same ten-task suite; ESI improves on these by 14.76 Score and 12.81 percentage points of SR.

\paragraph{Where gains appear} In sorting-sequence imitation, ESI achieves 94.12\% success, against 36\% for GPT-6-Astra and 0\% for $\pi_{0.5}$. In packing, the two component references score 11.60 and 19.10, respectively, and ESI reaches 65.75 with 50\% success. Language-conditioned classification shows a related pattern: ESI reaches 75\% success, compared with 30\% and 0\% for its components. These tasks require an object choice or sequence decision to be carried through several manipulation steps.

\paragraph{Progress versus completion} A higher Score can coexist with few complete episodes. On table organization, ESI reaches 63.75 Score but finishes only two of twenty cases. On pen placement, ESI improves Score over both references (33.25 versus 23.23 for $\pi_{0.5}$ and 5.00 for GPT-6-Astra), while its 5.00\% SR falls below the policy reference of 7.33\%. Sweeping also remains difficult, with two successes in twenty cases. These outcomes motivate examining which final task conditions remain unsatisfied after partial progress.

\subsection{Improvement over the matched baseline}

\begin{table}[!htbp]
\centering\small
\setlength{\tabcolsep}{3pt}
\renewcommand{\arraystretch}{1.2}
\caption{\textbf{Matched baseline--ESI comparison.} Main evaluation: ten tasks with twenty scheduled cases each. The last two columns compare the 41 cases completed by both agents. Baseline uses task-only conditioning. Interventions count teacher intervention decisions, and steps count teacher-executed actions per episode; step share pools executed steps. Changes in SR and share are percentage points; control-count changes are relative to baseline.}
\label{tab:baseline-comparison}\label{tab:matched-success}
\begin{tabularx}{\linewidth}{@{}l*{7}{C}@{}}
\toprule
& \multicolumn{5}{c}{All matched cases} & \multicolumn{2}{c}{Shared successes}\\
\cmidrule(lr){2-6}\cmidrule(l){7-8}
Agent & \shortstack{SR (\%)\\$\uparrow$} & \shortstack{Score\\$\uparrow$} & \shortstack{Interventions\\$\downarrow$} & \shortstack{Teacher\\steps $\downarrow$} & \shortstack{Step share\\(\%) $\downarrow$} & \shortstack{Interventions\\$\downarrow$} & \shortstack{Teacher\\steps $\downarrow$}\\
\midrule
\shortstack[l]{Baseline\\{\scriptsize(task-only)}} & 29.68 & 45.67 & 27.08 & 187.97 & 19.26 & 10.85 & 79.41\\
\textbf{ESI} & \textbf{49.41} & \textbf{62.03} & \textbf{15.92} & \textbf{100.28} & \textbf{11.12} & \textbf{4.71} & \textbf{29.76}\\
Change & +19.74 & +16.36 & $-41.20\%$ & $-46.65\%$ & $-8.13$ & $-56.63\%$ & $-62.53\%$\\
\bottomrule
\end{tabularx}
\end{table}

\begin{figure}[t]
\centering
\includegraphics[width=\linewidth]{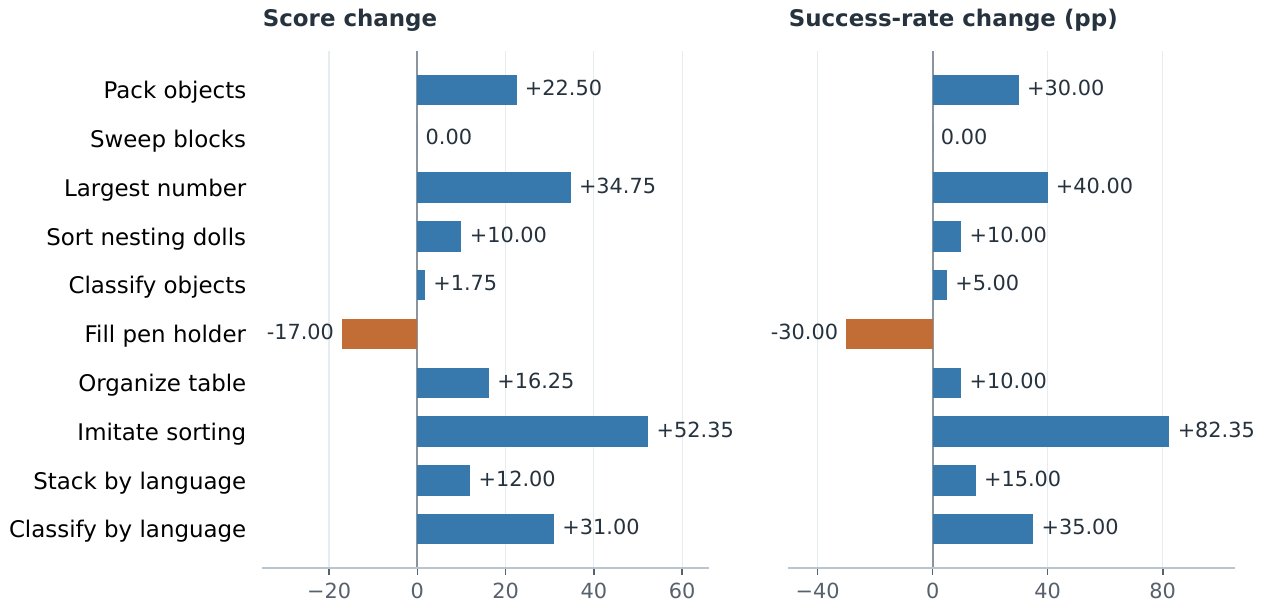}
\caption{\textbf{Task-level changes relative to the matched baseline.} Left: change in Score. Right: change in success rate, in percentage points. Blue denotes improvement, orange denotes decline, and zero indicates no aggregate change. Task order matches the main comparison table.}
\label{fig:task-gains}
\end{figure}

Table~\ref{tab:baseline-comparison} compares ESI with the original hybrid agent. Task-mean SR increases from 29.68\% to 49.41\%, a gain of 19.74 percentage points, and Score rises from 45.67 to 62.03, a gain of 16.36. ESI improves both metrics on eight tasks, matches the baseline on sweeping, and declines on pen placement (Figure~\ref{fig:task-gains}).

The largest gains occur in sorting-sequence imitation (SR 11.76\% to 94.12\%; Score 44.71 to 97.06), number arrangement (35\% to 75\%; 48.50 to 83.25), language-conditioned classification (40\% to 75\%; 50.50 to 81.50), and packing (20\% to 50\%; 43.25 to 65.75). Pen placement falls from 35\% to 5\% SR and from 50.25 to 33.25 Score, identifying a concrete priority for subsequent collection and refinement.

The matched outcomes show how the improvement is distributed across cases: 41 are completed by both systems, 55 change from baseline failure to ESI success, 18 change in the reverse direction, and 83 remain unsuccessful. ESI therefore completes 37 additional cases, increasing the total from 59 to 96.

\subsection{Control allocation and teacher intervention}

\paragraph{Fewer teacher-executed actions}
The teacher executes 187.97 action steps per episode in the baseline and 100.28 in ESI, a reduction of 46.65\%. Its share of pooled executed steps decreases from 19.26\% to 11.12\%, with VLA actions accounting for the remaining 88.88\% in ESI. These statistics describe a shift in the physical control allocation toward the action policy.

\paragraph{The same successful cases}
The shared-success columns of Table~\ref{tab:baseline-comparison} compares the 41 cases completed by both systems. Teacher-executed steps fall from 79.41 to 29.76 per episode, a reduction of 62.53\%. The reduction in direct intervention therefore also appears when both systems reach full task completion on the same instances.

\paragraph{Intervention frequency}
Teacher intervention decisions with nonzero executed edit or EEF actions decrease from 27.08 to 15.92 per episode, a reduction of 41.20\%. On the 41 shared-success cases, they decrease from 10.85 to 4.71, a reduction of 56.63\%. These results show higher task completion with fewer direct teacher interventions, both in frequency and in executed action steps.

\subsection{Component rollback on matched task instances}
\label{sec:rollback}

We evaluate the three adaptation targets in Figure~\ref{fig:automation} on ten tasks with ten matched cases each (100 cases in total). Table~\ref{tab:rollback} indicates the adapted components retained in each configuration. \emph{Baseline guidance} (Prompt rollback) removes additional task explanations, operational suggestions, and reference images while retaining the native instruction and tool protocol. \emph{Free-form subtasks} (Skill rollback) replaces canonical skill assembly with concise free-form instructions while preserving subtask conditioning. \emph{Initial VLA weights} (VLA rollback) restores the RoboDojo-adapted initial policy. The ESI reference uses the same 100 instances extracted from the main evaluation; reference and rollback implementation details are given in Appendix~\ref{app:additional-eval}.

\begin{table}[!htbp]
\centering
\small
\renewcommand{\arraystretch}{1.16}
\caption{\textbf{Rollback of the three ESI adaptation components.} Ten tasks with ten cases each. Checks retain the adapted components; dashes mark rollback. SR and Score are task means.}
\label{tab:rollback}
\begin{tabularx}{\linewidth}{@{}lCCCCC@{}}
\toprule
& \multicolumn{3}{c}{Adapted components} & \multicolumn{2}{c}{Task performance} \\
\cmidrule(lr){2-4}\cmidrule(l){5-6}
Configuration & Prompt & Skill & VLA & SR (\%) $\uparrow$ & Score $\uparrow$ \\
\midrule
\textbf{ESI} & {\scriptsize\faCheck} & {\scriptsize\faCheck} & {\scriptsize\faCheck} & \textbf{52.00} & \textbf{64.60} \\
Baseline guidance & --- & {\scriptsize\faCheck} & {\scriptsize\faCheck} & 41.00 & 53.90 \\
Free-form subtasks & {\scriptsize\faCheck} & --- & {\scriptsize\faCheck} & 50.00 & 62.75 \\
Initial VLA weights & {\scriptsize\faCheck} & {\scriptsize\faCheck} & --- & 29.00 & 50.49 \\
\bottomrule
\end{tabularx}
\par\smallskip{\footnotesize ESI uses the matched main-evaluation subset; rollback runs use the subsequent implementation. Evaluation definitions are given in Appendix~\ref{app:additional-eval}.}
\end{table}

\begin{figure}[!htb]
\centering
\includegraphics[width=\linewidth]{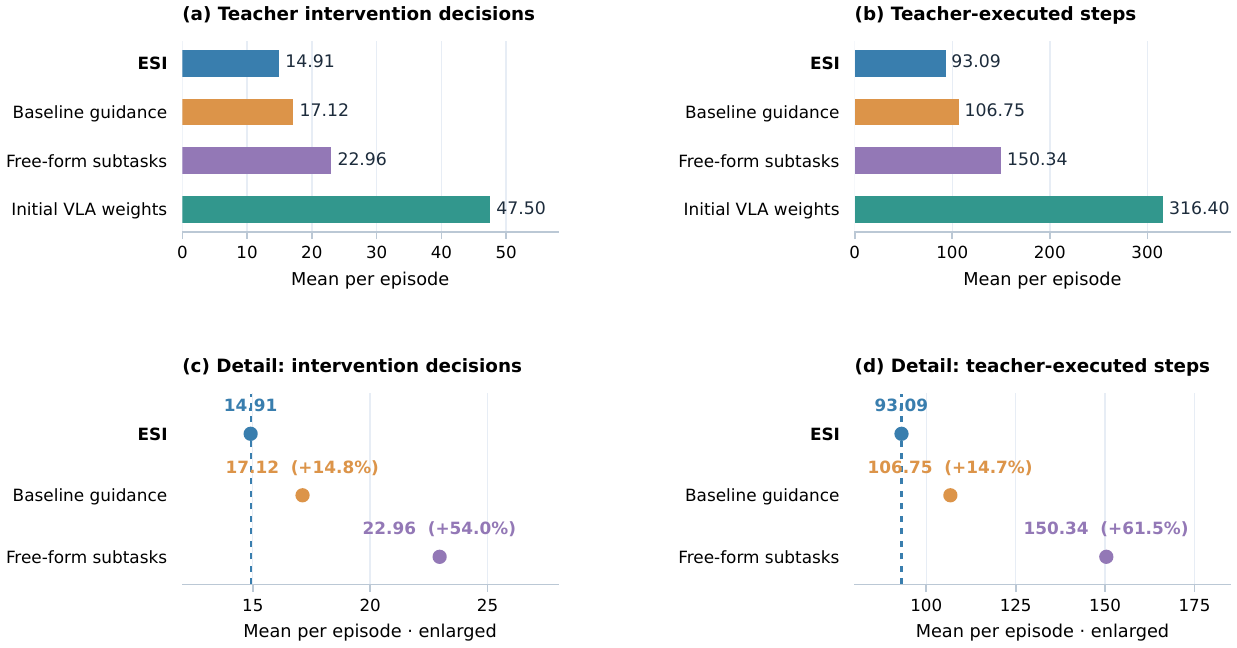}
\caption{\textbf{Teacher control under component rollback.} Top: all configurations on zero-based axes. Bottom: enlarged views of ESI, Baseline guidance, and Free-form subtasks; dashed lines mark ESI. Values are mean teacher intervention decisions or teacher-executed steps per episode; parentheses give relative differences from the matched ESI reference in Table~\ref{tab:rollback}.}
\label{fig:rollback-control}
\end{figure}

The matched ESI reference achieves 52.00\% SR and 64.60 Score. The \emph{Baseline guidance} configuration achieves 41.00\% SR and 53.90 Score, while the \emph{Initial VLA weights} configuration achieves 29.00\% SR and 50.49 Score. Figure~\ref{fig:rollback-control} shows the accompanying control allocation: the \emph{Initial VLA weights} configuration requires the most direct teacher control, with 47.50 teacher intervention decisions and 316.40 executed steps per episode. These results are consistent with the roles of task guidance in directing interaction and policy adaptation in enabling the student to carry out the selected operations.

The \emph{Free-form subtasks} configuration achieves 50.00\% SR and 62.75 Score, close to the ESI reference in task outcomes, while recording 22.96 teacher intervention decisions and 150.34 teacher-executed steps per episode. Instructions to produce concise subtasks remain active, and observed free-form descriptions are similar in intent and granularity to canonical skills. This offers a possible explanation for the retained task performance: the policy still receives familiar local goals despite the relaxed serialization. The free-form study further motivates separating the semantic content of an operation from its exact serialization. Relative to the matched ESI reference, \emph{Baseline guidance} records 14.8\% more teacher intervention decisions and 14.7\% more teacher-executed steps; \emph{Free-form subtasks} records 54.0\% and 61.5\% more, respectively (Figure~\ref{fig:rollback-control}, lower panels). These control measurements describe the teacher intervention accompanying each configuration. Together, task outcomes and direct control provide complementary views of the deployed collaboration. Task-level results appear in Appendix~\ref{app:additional-eval}.

\subsection{Transfer beyond the flywheel task suite}
\label{sec:transfer}

We evaluate \texttt{solve\_equation}, a task excluded from ESI's flywheel collection and post-training, on five paired cases. Both agents use subtask conditioning and action review. In this and the following order study, \emph{Baseline (subtask-conditioned)} denotes the agent with the RoboDojo-adapted initial policy, free-form subtasks, and no additional task guidance; ESI uses the adapted policy, canonical interface, and task guidance. The shorter label Baseline in the case-study figures refers to this subtask-conditioned agent. These case studies examine how the composed agent applies familiar operations to a new task or a different requested sequence. The evaluation protocol is given in Appendix~\ref{app:additional-eval}.

\begin{table}[!htbp]
\centering\small
\renewcommand{\arraystretch}{1.18}
\caption{\textbf{Task transfer and alternative-order execution.} Five paired cases of Solve equation and three requested orders per agent in one fixed scene. Both agents use subtask conditioning and action review. Baseline is subtask-conditioned. Interventions count teacher intervention decisions, and teacher steps count executed teacher actions, averaged per episode; teacher step share pools executed steps. Figures~\ref{fig:solve-case} and~\ref{fig:order-cases} visualize the two studies.}
\label{tab:transfer-order}
\begin{tabularx}{\linewidth}{@{}lYCCCC@{}}
\toprule
Experiment & Agent & Successes & Interventions & \shortstack{Teacher\\steps} & \shortstack{Teacher step\\share (\%)} \\
\midrule
Solve equation & Baseline & 2/5 & 15.00 & 99.00 & 42.45 \\
 & ESI & 4/5 & 7.40 & 43.60 & 21.52 \\
\midrule
Three orders & Baseline & 2/3 & 36.00 & 294.33 & 33.16 \\
 & ESI & 3/3 & 8.00 & 71.33 & 11.74 \\
\bottomrule
\end{tabularx}
\end{table}

ESI completes four of the five cases, compared with two for Baseline. Mean teacher intervention decisions decrease from 15.00 to 7.40, and teacher-executed steps decrease from 99.00 to 43.60, a reduction of 55.96\%. Both control measures decrease in every paired case (Table~\ref{tab:transfer-order}).

\begin{figure}[!tb]
\centering
\includegraphics[width=0.94\linewidth]{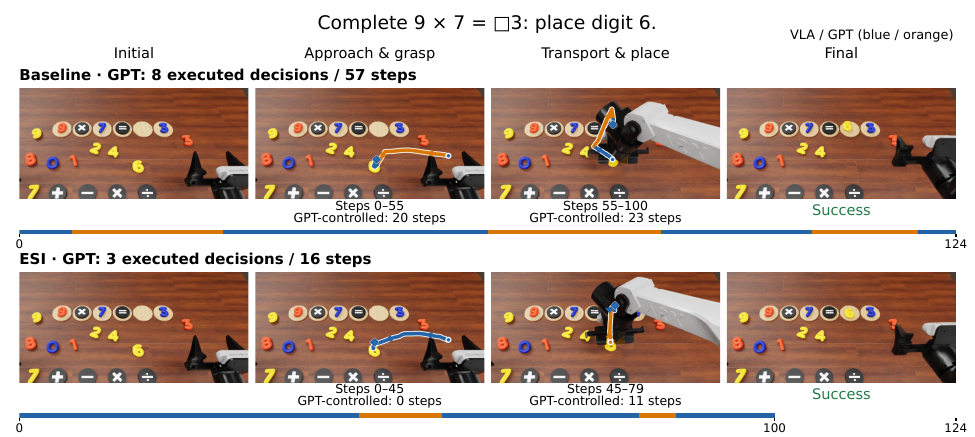}
\caption{\textbf{Executing a new task with less direct teacher control.} Both agents solve the same equation. Curves trace the projected gripper reference point; blue and orange denote VLA and teacher control. Strips show executed control on a shared step axis. GPT counts report teacher intervention decisions and teacher-executed steps; Baseline is subtask-conditioned.}
\label{fig:solve-case}
\end{figure}

Figure~\ref{fig:solve-case} follows a paired successful episode in which the missing digit in $9\times7=\square3$ is placed. ESI completes the approach-and-grasp phase with no GPT-controlled steps, compared with 20 for Baseline. Across the whole episode, direct teacher control decreases from eight decisions and 57 steps to three decisions and 16 steps. The teacher identifies the required digit and expresses the operation; the adapted policy supplies more of the resulting motion.

\subsection{Intent following under reordered operations}
\label{sec:order}

We test whether the agent can reorganize learned operations according to language-specified intent, rather than replay a fixed execution sequence associated with a familiar scene. The initial scene and final goal remain unchanged, while the requested order of the placements varies.

We evaluate three requested operation orders in one fixed initial scene of \texttt{arrange\_largest\_number}, with three runs per agent. The pad-filling orders are: $1\!\rightarrow\!2\!\rightarrow\!3\!\rightarrow\!4$, $2\!\rightarrow\!3\!\rightarrow\!1\!\rightarrow\!4$, and $4\!\rightarrow\!3\!\rightarrow\!2\!\rightarrow\!1$, with pads numbered from left to right. The target arrangement remains $8720$. Video inspection verifies adherence to the requested order; native scoring separately measures task completion. ESI completes the task under all three requested orders, while Baseline completes it under two (Table~\ref{tab:transfer-order}).

\begin{figure}[!tb]
\centering
\includegraphics[width=0.92\linewidth]{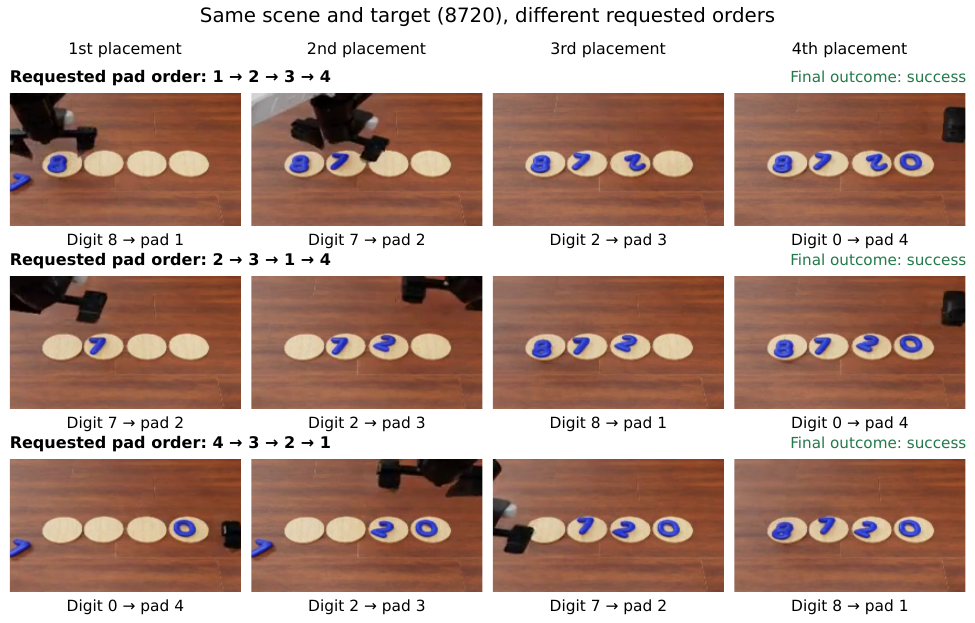}
\caption{\textbf{Language-conditioned intent following in a fixed scene.} Rows show three requested pad-filling orders toward the same target, $8720$. Columns show the first placement on each successive pad, numbered left to right; labels identify the digit and destination. All panels use the same scene crop. Final outcomes include subsequent pose adjustments.}
\label{fig:order-cases}
\end{figure}

Figure~\ref{fig:order-cases} makes the dependence of execution on intent visible. Its rows share the same scene and target arrangement, while each column advances one requested placement. ESI first places digit 8, 7, or 0 when the requested order begins with pad 1, 2, or 4, respectively, and continues with the remaining placements in the specified order. These sequences support compositional intent following: familiar local operations can be selected and reordered through language, extending their use beyond one scene-associated execution sequence.

Across these runs, mean teacher intervention decisions fall from 36.00 to 8.00 and teacher-executed steps from 294.33 to 71.33, reductions of 77.78\% and 75.76\%, respectively. Teacher step share decreases from 33.16\% to 11.74\%. Together with the observed order adherence, reduced direct teacher control supports the view that the adapted VLA carries more of the execution needed to realize changing local intentions. This illustrates the behavioral role of the shared LLM--VLA interface: the teacher organizes the task through local instructions, and the policy supplies the corresponding motion. The order study complements the held-out-task example by changing the intended sequence rather than the task scene. Appendix~\ref{app:additional-eval} includes per-order outcomes and token measurements under the shared subtask-conditioning control flow.

\section{Conclusion and Outlook}

ESI brings system-level post-training to modular embodied agents through an automated self-improving flywheel. It updates the prompts that guide the LLM, the skill catalog that expresses intent, and the VLA parameters that realize it. Through a shared control and learning interface, this process transfers semantic intent and corrective behavior into persistent action capability. The updated components are deployed together for the next round.

On ten selected RoboDojo tasks, the evaluated ESI agent achieves 62.03 mean Score and 49.41\% mean success rate, exceeding both standalone component references. Against the matched hybrid baseline, task-mean success improves by 19.74 percentage points and teacher-executed steps per episode decrease by 46.65\%. Transfer and alternative-order case studies extend this picture: the adapted agent completes four of five held-out-task cases and all three requested order variants with reduced teacher control. ESI establishes a practical direction for embodied systems in which the experience of cooperating becomes a resource for improving future cooperation, and a reusable pathway for translating foundation-model advances into robot capabilities.

\subsection{Research directions}
\label{sec:roadmap}

\paragraph{Foundation-model upgrades}
This perspective opens a second axis of improvement alongside additional robot experience: upgrading the reasoning foundation model. A more capable teacher can supply new operation decompositions, richer scene interpretations, and useful recovery strategies for the system to explore and absorb. The shared interface and automated learning loop provide the means to translate these contributions into policy capability. The research opportunity is to make progress in general-purpose intelligence a continuing source of progress in robot execution.

The framework supports examining both immediate and persistent benefits of a stronger teacher. Immediate benefits appear when the teacher is changed during deployment. Persistent benefits can be studied by collecting experience with different teachers, training the resulting students under matched budgets, and evaluating those students with one common teacher. This separates improvements in online guidance from the capabilities retained through learning and offers a practical way to investigate foundation-model-driven embodied learning.

\paragraph{Interaction-driven learning}
The teacher encounters the student's current weaknesses during deployment. As the policy learns an operation, subsequent interaction can expose later-stage failures or more demanding combinations of skills. This creates an opportunity to study curricula driven by the evolving boundary of the agent's capability. Experience-based collection priorities could direct additional effort toward recurring difficulties and useful recoveries.

Future experiments can hold the teacher's instruction stream fixed to isolate policy-level intent following, or start policies from identical corrected states to examine continuation after intervention.

A correction can contribute both to the current episode and to future policy capability. This motivates tracking task performance together with intervention burden and interaction cost across rounds. Policy adaptation and review scheduling offer complementary research directions: one changes the actions the student proposes, while the other determines when additional reasoning is most useful.

\paragraph{An extensible framework}
The prompt, skill interface, and policy provide distinct points for extending ESI. Different reasoning models can review experience, alternative skill representations can express intent, and other action models or training objectives can absorb effective behavior. Their common connection is the interaction record, which links decisions to actions and outcomes. The framework supports studying how these choices affect the resulting embodied agent.


\clearpage
\bibliographystyle{IEEEtranN}
\bibliography{references}

\appendix

\section{Shared Subtask Interface}
\label{app:schema}

Table~\ref{tab:schema} summarizes the interface used by the online teacher and offline data pipeline. Entity references use stable snake-case names and are rendered as readable text. Unused optional fields remain null or take their designated neutral enum value. An unresolved operation can be represented by a null subtask, preserving task-only conditioning.

\begin{table}[H]
  \captionsetup{justification=raggedright,singlelinecheck=false}
  \caption{Fields of the shared subtask representation.}
  \label{tab:schema}
  \centering\small
  \begin{tabularx}{\linewidth}{@{}p{0.31\linewidth}Y@{}}
    \toprule
    \tablehead{Field} & \tablehead{Meaning} \\
    \midrule
    \texttt{skill\_type} & Standard operation name or \texttt{custom}. \\
    \texttt{arm} & Acting arm: \texttt{left}, \texttt{right}, or \texttt{both}. \\
    \texttt{object\_ref} & Object or entity involved in the operation. \\
    \texttt{destination\_ref} & Target entity or region for operations that require one. \\
    \texttt{target\_relation} & \texttt{none}, \texttt{to}, \texttt{on}, \texttt{in}, or \texttt{near}. \\
    \texttt{target\_orientation\_ref} & Desired orientation of the object. \\
    \texttt{custom\_description} & Concise free-form operation used only for \texttt{custom}. \\
    \texttt{giver\_arm}, \texttt{receiver\_arm} & Arm identities for handover. \\
    \bottomrule
  \end{tabularx}
\end{table}

Representative renderer outputs include:
\begin{promptbox}[Canonical subtask examples]
Left arm: Grasp red block.\\
Right arm: Place blue cup on tray.\\
Left arm: Rotate pen to tip downward.\\
Transfer red block from the left gripper to the right gripper.\\
Both arms: Return home.
\end{promptbox}

The orientation field describes the desired object orientation. The teacher and policy still need to determine the physical motion that realizes it. The schema supplies an intent representation rather than a geometric planner.

\section{Update Procedures and Policy Training Details}
\label{app:config}

This appendix specifies the review procedure and data-processing rules used by the updates in Section~\ref{sec:learning}, followed by the reported policy-training configurations.

\subsection{Prompt review and a recorded proposal}
\label{app:prompt-review}

An independent reviewer checks candidate guidance against the cited episode summaries, reasoning, feedback, and images, including successful continuations that a new rule should preserve. The review examines whether the evidence supports the proposed failure mechanism, whether the trigger can be observed online, and whether the revision preserves task goals and the execution protocol. It uses historical interaction records without launching additional rollouts. Revisions that need further evidence return for refinement. The complete guidance bundle is automatically published after it passes review and replaces the corresponding prompt content in the next collection round.

\Needspace{15\baselineskip}
\begin{promptbox}[Recorded example: a keyboard-contact guidance proposal]
\raggedright
\textbf{Existing guidance.} Push in short segments, reassess displacement and rotation, and stop repeating a blocked approach.\\[3pt]
\textbf{Interaction evidence.} During table organization, the recorded review observes the fingertips crossing the keyboard border without visible keyboard displacement. Other inspected segments contain limited progress toward the required rotation.\\[3pt]
\textbf{Proposed refinement.} Separate missed edge engagement from contact that produces the wrong motion: adjust approach height in the former case and contact location in the latter, while preserving productive student pushes.\\[3pt]
\textbf{Review outcome.} The independent reviewer found that contact and height adjustments already occurred in the inspected execution, and requested stronger evidence connecting the difficulty to missing guidance. This candidate remained unpublished.
\end{promptbox}

The example shows how a broad operational instruction becomes a concrete, observation-conditioned proposal and how historical evidence determines whether that proposal enters the next guidance bundle.

\subsection{Annotation and local eligibility}
\label{app:annotation}

Each executed command is aligned with its preceding observation. An LLM annotator uses the textual interaction record and post-action feedback to label local outcomes. For the first collection round, which precedes online subtask conditioning, offline annotation groups decisions into operation attempts and assigns subtask instructions. For later rounds, annotation preserves the original online instructions and executed decision intervals, adding outcome and instruction--action consistency labels.

Supported successful execution is retained for both student actions and teacher corrections, including preparatory actions whose useful progress is verified by subsequent feedback. Failed, uncertain, or inconsistent segments are excluded from direct action supervision. Eligibility is independent of the episode's final outcome: a completed grasp and transport can be retained even if a later placement fails, while an unsuccessful attempt within a successful episode can be excluded. Across the three collection rounds, 47.9--49.7\% of candidate training-window starts come from episodes that ultimately fail to complete the task.

\subsection{Window construction and validity masks}
\label{app:windows}

Eligible execution is organized into contiguous intervals, from which the pipeline extracts overlapping action windows of horizon $H=50$. A change between student control and teacher correction does not break an interval. Full-length windows provide the main training samples. Additional tail windows are retained when teacher control occurs near the end of a valid interval, provided at least 15 valid action steps remain.

For a selected start $t$, let $L_t\leq H$ be the length of its contiguous valid prefix. The temporal mask $m_{t,k}$ is one for $k<L_t$ and zero otherwise. Supervision stops at the first ineligible position or episode end and does not resume if later positions become eligible again. Invalid and padded positions therefore receive no direct action-loss supervision.

\subsection{Language conditioning at operation transitions}
\label{app:conditioning}

Each window uses the observation and one language condition at its starting point; the condition is not reassigned at every future action. The loader selects between the local subtask and the episode goal according to the following cutoffs.

For online-conditioned data, let $[a,b)$ be the final positive-length decision in a contiguous run of identical subtask text. The loader retains the subtask for window starts $t<a+\lceil(b-a)/2\rceil$ and otherwise uses the episode goal alone. For an offline-labeled first-round segment $[s,e)$, it retains the subtask when $e-t\geq25$ if the segment has at least 50 steps, or when $2(t-s)<e-s$ for shorter segments. These rules select one condition per window independently of the action-loss mask. Figure~\ref{fig:training-example} illustrates the online rule with two indexed training windows.

A controller switch preserves the action sequence, a subtask boundary affects the window's language selection, and a validity boundary terminates direct action supervision. These boundaries have distinct roles and need not coincide.

\subsection{A recorded interaction-to-training example}
\label{app:training-example}

\begin{figure}[!htbp]
\centering
\includegraphics[width=\linewidth]{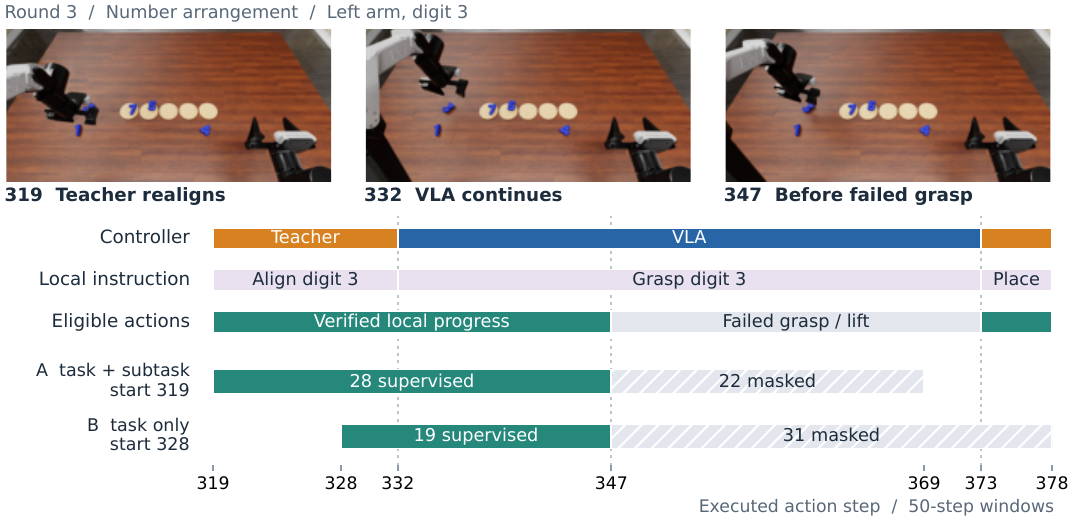}
\caption{\textbf{From a recorded interaction to policy supervision.} Three frames from a third-round episode align with the controller, instruction, and eligibility tracks. Control and instruction change at 332; eligibility ends at 347. The loader gives window A (start 319) the alignment subtask and window B (start 328) task-only conditioning, using the midpoint cutoff of decision $[324,332)$. Both 50-step masks stop at 347, even when eligibility resumes at 373. Step intervals are zero-based and half-open.}
\label{fig:training-example}
\end{figure}

Figure~\ref{fig:training-example} follows a recorded third-round number-arrangement episode. The teacher realigns the left hand with digit 3 during steps $[319,332)$; the VLA then makes a useful preparatory adjustment during $[332,347)$. Subsequent grasp and lift attempts are labeled unsuccessful, so the valid interval ends at step 347. A window starting at 319 retains the local instruction ``Left arm: Align with digit 3.'' and supervises 28 actions across the controller handoff, masking the following 22 positions. A window starting at 328 uses the episode goal alone and supervises 19 actions, with 31 masked positions. These windows share the same validity boundary but use different language conditions. The episode does not complete the task, yet its verified correction and continuation remain usable supervision.

\subsection{Replay weights and within-round sampling}
\label{app:replay}

\paragraph{Recursive round weights}
Under the recursive rule in Equation~\ref{eq:replay}, the sampling mass assigned to data from round $i$ at training round $r$ is
\begin{equation}
w_i^{(r)}=
\begin{cases}
(1-\alpha)^{r-1}, & i=1,\\
\alpha(1-\alpha)^{r-i}, & 2\leq i\leq r.
\end{cases}
\label{eq:round-weights}
\end{equation}
With $\alpha=0.6$, the first four mixtures, ordered from oldest to newest, are $(1)$, $(0.4,0.6)$, $(0.16,0.24,0.60)$, and $(0.064,0.096,0.24,0.60)$. Thus, the 40\% historical share preserves the same recursive allocation at every additional round.

\paragraph{Within-round sampling}
Within each round, sampling assigns equal mass to tasks and balances the available standard/random splits. Eligible cases receive relative weights of 2:1 according to whether their episodes finish successfully or unsuccessfully, and repeated episodes within a case and outcome group share its mass equally. Within an episode, teacher-controlled window starts receive weight 4 and student-controlled starts weight 1, normalized over that episode's candidate starts. Case-level allocation prevents long episodes from dominating merely by providing more windows, while source weighting increases the opportunity to learn from corrective execution. These factors define $\mu_r$; their effect on training enters through sample selection.

\subsection{Masked flow-matching objective}
\label{app:objective}

For a window start $t$ sampled from the replay distribution $q_r$, let $\ell^{\mathrm{FM}}_{t,k}(\theta)$ denote the native flow-matching action loss at position $k$. With the temporal mask from Appendix~\ref{app:windows}, the training loss is
\begin{equation}
\mathcal{L}_r(\theta)=\mathbb{E}_{t\sim q_r}\left[
\frac{\sum_{k=0}^{H-1}m_{t,k}\ell^{\mathrm{FM}}_{t,k}(\theta)}{\sum_{k=0}^{H-1}m_{t,k}}
\right].
\label{eq:loss}
\end{equation}
Normalizing by the number of valid positions gives each sampled window equal overall weight, including retained short tails.

\subsection{Training configuration}
\label{app:training-config}

Table~\ref{tab:config} summarizes the third-round policy training configuration.

\begin{table}[H]
  \captionsetup{justification=raggedright,singlelinecheck=false}
  \caption{Third-round configuration snapshot.}
  \label{tab:config}
  \centering\small
  \begin{tabularx}{\linewidth}{@{}p{0.49\linewidth}Y@{}}
    \toprule
    \tablehead{Setting} & \tablehead{Configured value} \\
    \midrule
    Student initialization & Round-two checkpoint at step 6000 \\
    Language mode & Task plus current subtask, with transition fallback \\
    Action horizon / minimum retained tail & 50 / 15 steps \\
    Round-one / two / three sampling mass & 0.16 / 0.24 / 0.60 \\
    Task and standard/random balancing & Uniform task mass; 0.5 / 0.5 variant mass \\
    Successful / unsuccessful case weights & 2 / 1 \\
    Teacher / student window-start weights & 4 / 1 \\
    Optimization updates & 6000 \\
    Global batch / global microbatch & 256 / 128 \\
    Peak / terminal learning rate & $2\times10^{-5}$ / $10^{-6}$ \\
    Warmup & 600 updates \\
    Device count / FSDP devices & 16 / 8 \\
    Gradient norm clipping & 1.0 \\
    \bottomrule
  \end{tabularx}
\end{table}

The ten tasks cover packing objects into a box, arranging the largest number, sorting nesting dolls by size, classifying objects, filling a pen holder, organizing a table, imitating a sorting sequence, stacking blocks by language, classifying objects by language, and sweeping blocks. The suite specifies 20 cases per task in each round.

\section{Additional Evaluation Details and Task-Level Results}
\label{app:additional-eval}

\paragraph{Cases and configuration}
The component study uses ten cases per task under the sampling rule in Section~\ref{sec:results}. All 100 instances match the ESI reference by task, split, and layout. The reference is extracted from the earlier main evaluation; rollback runs use the subsequent frozen evaluation implementation. Their absolute results are reported without assigning the cross-version differences to individual components. The main evaluation uses execution configuration \texttt{joint\_fk\_rdp\_significant\_gripper\_v2}; subsequent rollback and paired case studies share their updated execution configuration, recorded in the archived run configurations. The paired transfer and order comparison preserves subtask conditioning in both agents and differs from the task-only main baseline.

\paragraph{Outcome and control aggregation}
The main comparison excludes three native-invalid layouts identically from both agents, leaving 197 paired scored cases. Scoreable failures remain included. Rollback SR counts success over all ten planned cases per task, with the three-hour timeout counted as non-success. Score averages available native scores within each task and then weights tasks equally. The policy-initialization rollback has 80 native scores and 20 timeouts; its task-mean Score is 50.49. Control means use the executed records of all selected cases through termination or timeout. A teacher intervention decision executes at least one edit or EEF action. Teacher step share is the sum of teacher-executed steps divided by the sum of all executed environment steps. Teacher intervention decision share uses positive-step teacher intervention decisions divided by all action-review decisions; subtask-planning decisions are excluded from this denominator. Shares pool counts rather than average episode percentages.

\paragraph{Transfer and order protocol}
The reported Solve layouts 45--49 were selected for a follow-up case study after preliminary testing on layouts 0--4. ``Held out'' refers to ESI flywheel collection and post-training. Each order condition uses standard layout 0 and a distinct requested pad-filling sequence. Video inspection confirms order adherence in the executed placements; native completion is a separate outcome. The ESI 2314 condition uses a completed replacement run after an initial timeout, retained in the attempt archive. Token statistics below describe the selected runs, including observed within-run retry usage, rather than all attempted runs.

\begin{table}[!htbp]
\centering\small
\renewcommand{\arraystretch}{1.16}
\caption{\textbf{Task-level component results.} Cells show Score / SR (\%). Each task has ten planned cases. The reference and outcome definitions follow Table~\ref{tab:rollback}.}
\label{tab:rollback-tasks}
\begin{tabularx}{\linewidth}{@{}lCCCC@{}}
\toprule
Task & ESI reference &  \shortstack{Baseline\\guidance} & \shortstack{Free-form\\subtasks} & \shortstack{Initial VLA\\weights} \\
\midrule
Pack objects & 64.50 / 50\% & 35.00 / 10\% & 50.00 / 30\% & 30.83 / 10\% \\
Sweep blocks & 10.00 / 10\% & 10.00 / 10\% & 10.00 / 10\% & 25.00 / 20\% \\
Largest number & 94.00 / 90\% & 66.50 / 50\% & 71.50 / 60\% & 67.14 / 40\% \\
Sort nesting dolls & 30.00 / 30\% & 10.00 / 10\% & 60.00 / 60\% & 37.50 / 30\% \\
Classify objects & 85.50 / 80\% & 94.00 / 90\% & 88.00 / 80\% & 86.67 / 70\% \\
Fill pen holder & 32.50 / 0\% & 46.00 / 20\% & 34.50 / 10\% & 52.22 / 20\% \\
Organize table & 67.50 / 10\% & 47.50 / 0\% & 62.50 / 20\% & 53.12 / 0\% \\
Imitate sorting & 100.00 / 100\% & 75.00 / 70\% & 88.00 / 80\% & 50.00 / 20\% \\
Stack by language & 84.00 / 80\% & 64.00 / 60\% & 84.00 / 80\% & 68.00 / 60\% \\
Classify by language & 78.00 / 70\% & 91.00 / 90\% & 79.00 / 70\% & 34.44 / 20\% \\
\midrule
Task mean & 64.60 / 52.00\% & 53.90 / 41.00\% & 62.75 / 50.00\% & 50.49 / 29.00\% \\
\bottomrule
\end{tabularx}
\end{table}

\begin{table}[H]
\centering\small
\renewcommand{\arraystretch}{1.15}
\caption{\textbf{Order-conditioned case results.} A fixed initial layout is evaluated under three requested filling sequences. Video inspection confirms order adherence; completion is measured by the native task evaluator. Baseline is subtask-conditioned; interventions and teacher steps follow the definitions in Section~\ref{sec:results}.}
\label{tab:order-details}
\begin{tabular*}{\linewidth}{@{\extracolsep{\fill}}llrrrr@{}}
\toprule
Order & Agent & Complete & Score & Interventions & Teacher steps \\
\midrule
1234 & Baseline & No & 30.00 & 41 & 366 \\
1234 & ESI & Yes & 100.00 & 11 & 138 \\
\midrule
2314 & Baseline & Yes & 100.00 & 43 & 368 \\
2314 & ESI & Yes & 100.00 & 7 & 43 \\
\midrule
4321 & Baseline & Yes & 100.00 & 24 & 149 \\
4321 & ESI & Yes & 100.00 & 6 & 33 \\
\bottomrule
\end{tabular*}
\end{table}

\begin{table}[H]
\centering\small
\renewcommand{\arraystretch}{1.15}
\caption{\textbf{Token usage in the order case study.} Baseline is subtask-conditioned. Means per selected episode under the shared subtask-conditioning and action-review flow. K and M denote thousands and millions. Total includes cached input; output already includes reasoning tokens.}
\label{tab:order-tokens}
\begin{tabular*}{\linewidth}{@{\extracolsep{\fill}}lrrr@{}}
\toprule
Token category & Baseline & ESI & Reduction \\
\midrule
Uncached input (K) & 435.62 & 317.73 & 27.06\% \\
Cached input (M) & 12.243 & 7.322 & 40.19\% \\
Output (K) & 59.36 & 29.61 & 50.13\% \\
Total (M) & 12.738 & 7.669 & 39.79\% \\
\bottomrule
\end{tabular*}
\par\smallskip{\footnotesize These figures describe the selected runs, not the aggregate cost of all attempts. No subscription-credit or monetary conversion is applied.}
\end{table}

\end{document}